\documentclass{article}

\usepackage[preprint]{neurips_2020}

\usepackage[utf8]{inputenc} 
\usepackage[T1]{fontenc}    
\usepackage{hyperref}       
\usepackage{url}            
\usepackage{booktabs}       
\usepackage{amsfonts}       
\usepackage{nicefrac}       
\usepackage{microtype}      
\usepackage{graphicx}
\usepackage{wrapfig}
\usepackage{ctable}
\usepackage{amsmath}
\usepackage{amsthm}
\usepackage{enumitem}
\usepackage{multirow}
\usepackage{hhline}
\newtheorem{theorem}{Theorem}
\newtheorem{lemma}{Lemma}

\newcommand{\eg}{\emph{e.g.}}

\title{SOLAR: Sparse Orthogonal Learned and Random Embeddings}

\author{%
  Tharun Medini \\
  Electrical and Computer Engineering\\
  Rice University\\
  \texttt{tharun.medini@rice.edu} \\
    \And
    Beidi Chen \\
    Department of Computer Science \\
    Rice University \\
    \texttt{beidi.chen@rice.edu} \\
    \And
    Anshumali Shrivastava \\
    Department of Computer Science \\
    Rice University \\
    \texttt{anshumali@rice.edu} \\
}

\begin{document}

\maketitle
\begin{abstract}
Dense embedding models are commonly deployed in commercial search engines, wherein all the document vectors are pre-computed, and near-neighbor search (NNS) is performed with the query vector to find relevant documents. However, the bottleneck of indexing a large number of dense vectors and performing an NNS hurts the query time and accuracy of these models. In this paper, we argue that high-dimensional and ultra-sparse embedding is a significantly superior alternative to dense low-dimensional embedding for both query efficiency and accuracy. Extreme sparsity eliminates the need for NNS by replacing them with simple lookups, while its high dimensionality ensures that the embeddings are informative even when sparse. However, learning extremely high dimensional embeddings leads to blow up in the model size. To make the training feasible, we propose a partitioning algorithm that learns such high dimensional embeddings across multiple GPUs without any communication. This is facilitated by our novel asymmetric mixture of {\bf S}parse, {\bf O}rthogonal, {\bf L}earned {\bf a}nd {\bf R}andom (SOLAR) Embeddings. The label vectors are random, sparse, and near-orthogonal by design, while the query vectors are learned and sparse. We theoretically prove that our way of one-sided learning is equivalent to learning both query and label embeddings. With these unique properties, we can successfully train 500K dimensional SOLAR embeddings for the tasks of searching through 1.6M books and multi-label classification on the three largest public datasets. We achieve superior precision and recall compared to the respective state-of-the-art baselines for each task with up to ${\bf 10} \times$ faster speed.
\end{abstract}
\vspace{-0.4cm}
\section{Introduction}\label{sec:intro}
\vspace{-0.2cm}
Embedding models have been the mainstay algorithms for several machine learning applications like Information Retrieval (IR)~\citep{croft2010search,baeza1999modern} and Natural Language Processing (NLP)~\citep{word2vec, glove, attention, bert} in the last decade. Embedding models are learned spin-offs from the low-rank approximation and Matrix Factorization techniques that dominated the space of recommendation systems prior to the emergence of Deep Learning (DL). The primary purpose of these models is to project a rather simple and intuitive representation of an input to an abstract low-dimensional dense vector space. This projection enables two things: 1) tailoring the vectors to specific downstream applications and 2) pre-processing and storing documents or products as vectors, thereby making the retrieval process computationally efficient (often matrix multiplication followed by sorting, which are conducive to modern hardware like GPUs). 

Besides the computational advantage, embedding models capture the semantic relationship between queries and products. A good example is product prediction for a service like Amazon. A user-typed query has to be matched against millions of products and the best search results have to be displayed within a fraction of a second. With naive product data, it would be impossible to figure out that products with `aqua' in their titles are actually relevant to the query `water'. Rather, if we can project all the products to a dense low-dimensional vector space, a query can also be projected to the same space and an inner product computation can be performed with all the product vectors (usually a dot product). We can then display the products with the highest inner product. These projections can be learned to encapsulate semantic information and can be continually updated to reflect temporal changes in customer preference. To the best of our knowledge, embedding models are the most prevalent ones in the industry, particularly for product and advertisement recommendations (Amazon's - DSSM~\citep{dssm}, Facebook's DLRM~\citep{dlrm}).

However, the scale of these problems has blown out of proportion in the past few years prompting research in extreme classification tasks, where the number of classes runs into several million. Consequentially, approaches like Tree-based Models~\citep{prabhu2014fastxml,Pfastre,agrawal2013multi} and Sparse-linear Models~\citep{weston2013label,pdsparse,ppdsparse} have emerged as powerful alternatives. Particularly, Tree-based models are much faster to train and evaluate compared to the other methods. However, most real Information Retrieval systems have dynamically changing output classes and all the extreme classification models fail to generalize to  new classes with limited training data (\eg, new products being added to the catalogue every day). This has caused the resurgence of embedding models for large scale Extreme Classification~\citep{bi2013efficient,annexml,sleec,chen2012feature}.

{\bf Our Contributions:} In this paper, we make a novel, unique and powerful argument that sparse high dimensional embeddings are superior to their dense low dimensional counterparts. In this regard, we make two interesting design choices: 1) We design the label embeddings (\eg products in the catalogue) to be high dimensional, super-sparse, and orthogonal vectors. 2) We fix the label embeddings throughout the training process and learn only the input embeddings (one-sided learning), unlike typical dense models, where both the input and label embeddings are learned. Since we use a combination of {\bf S}parse, {\bf O}rthogonal, {\bf L}earned {\bf a}nd {\bf R}andom embeddings, we code-name our method {\bf SOLAR}. We provide a theoretical premise for SOLAR by showing that one-sided and two-sided learning are mathematically equivalent. Our choices manifest in a four-fold advantage over prior methods:
\begin{itemize}[leftmargin=*,nosep,nolistsep]
\itemsep0em
    \item {\bf Matrix Multiplication to Inverted-Index Lookup:} 
Sparse high dimensional embeddings can obtain a subset of labels using a mere inverted-index~\citep{croft2010search} lookup and restrict the computation and sorting to those labels. This enhances the inference speed by a large margin.
    \item {\bf Load-balanced Inverted Index:} By forcing the label embeddings to be near-orthogonal and equally sparse (and fixing them), we ensure that all buckets in an inverted index are equally filled and we sample approximately the same number of labels for each input. This omits the well-known imbalanced buckets issue where we sub-sample almost all the labels for popular inputs and end up hurting the inference speed.
    \item {\bf Learning to Hash:} An Inverted-Index can be perceived as a hash table where all the output classes are hashed into a few buckets~\citep{kulis2009learning, wang2017survey}. By fixing the label buckets and learning to map the inputs to the corresponding label buckets, we are doing a `partial learning to hash' task in the hindsight (more on this in Appendix A). 
    \item {\bf Zero-communication:} Our unique construction of label embeddings enables distributed training over multiple GPUs with zero-communication. 
Hence, we can afford to train on a 1.67 M book recommendation dataset and three largest extreme classification datasets and outperform the respective baselines on all 4 of them on both precision and speed.
\end{itemize}

\vspace{-0.5cm}
\section{Related Work}\label{sec:rel_work}
\vspace{-0.3cm}
{\bf SNRM:} While there have been a plethora of dense embedding models, there is only one prior work called SNRM (Standalone Neural Ranking Model)~\citep{snrm} that trains sparse embeddings for the task of suggesting documents relevant to an input query (classic web search problem). In SNRM, the authors propose to learn a high dimensional output layer and sparsify it using a typical L1 or L2 regularizer. 
However, imposing sparsity through regularization has multiple issues - 1) A large regularization weight causes the embeddings to be too sparse and we do not retrieve any labels for most inputs. On the other hand, a small regularization weight will end up retrieving too many candidates defeating the purpose of sparse embeddings. 2) The inverted-index generally has a lopsided label distribution causing imbalanced loads and high inference times. As we see in our experiments later, these issues lead to the poor performance of SNRM on our 1.67M product recommendation dataset.

{\bf GLaS:} Akin to SOLAR's construction of near-orthogonal label embeddings, another recent work from Google~\citep{glas} also explores the idea of enforcing orthogonality to make the labels distinguishable and thereby easier for the classifier to learn. The authors enforce it in such a way that frequently co-occurring labels have high cosine-similarity and the ones that rarely co-occur have low cosine similarity. This imposition was called a {\bf G}raph {\bf L}aplacian {\bf a}nd {\bf S}preadout ({\bf GLaS}) regularizer.  However, this was done entirely in the context of dense embeddings and cannot be extended to our case due to the differentiability issue. We show the comparison of SOLAR against dense embedding models with and without GLaS regularizer later on in section \ref{subsec:p2p}.

All other embedding models~\citep{bi2013efficient,annexml,sleec,chen2012feature} have primarily the same workflow of projecting inputs and respective labels to the same vector space and optimizing a similarity-based loss function. They differ in the choice of projection functions (Linear Projections vs Deep Neural Networks) and construction of Nearest Neighbour graphs for faster inference~\citep{annexml}. Including SNRM, all the embedding models have some common traits: 1) they train a pairwise loss function which means that every query-label pair is a separate training instance. This blows up the training data size and leads to long training times. 2) In order to avoid the degenerate case of all embeddings being the same, these models employ negative sampling~\citep{gutmann2010noise} techniques to identify irrelevant labels to the input and penalize the cosine similarity of the input-output embeddings. This exacerbates the data size even further.

SOLAR, in addition to being sparse, also solves these challenges by learning a classifier instead of a similarity based loss function, encapsulating all labels of an input at once. Since a classifier has intrinsic negative sampling, the number of training data samples is much lower.

\begin{figure}[!t]
    \centering
    \includegraphics[width=0.95\textwidth]{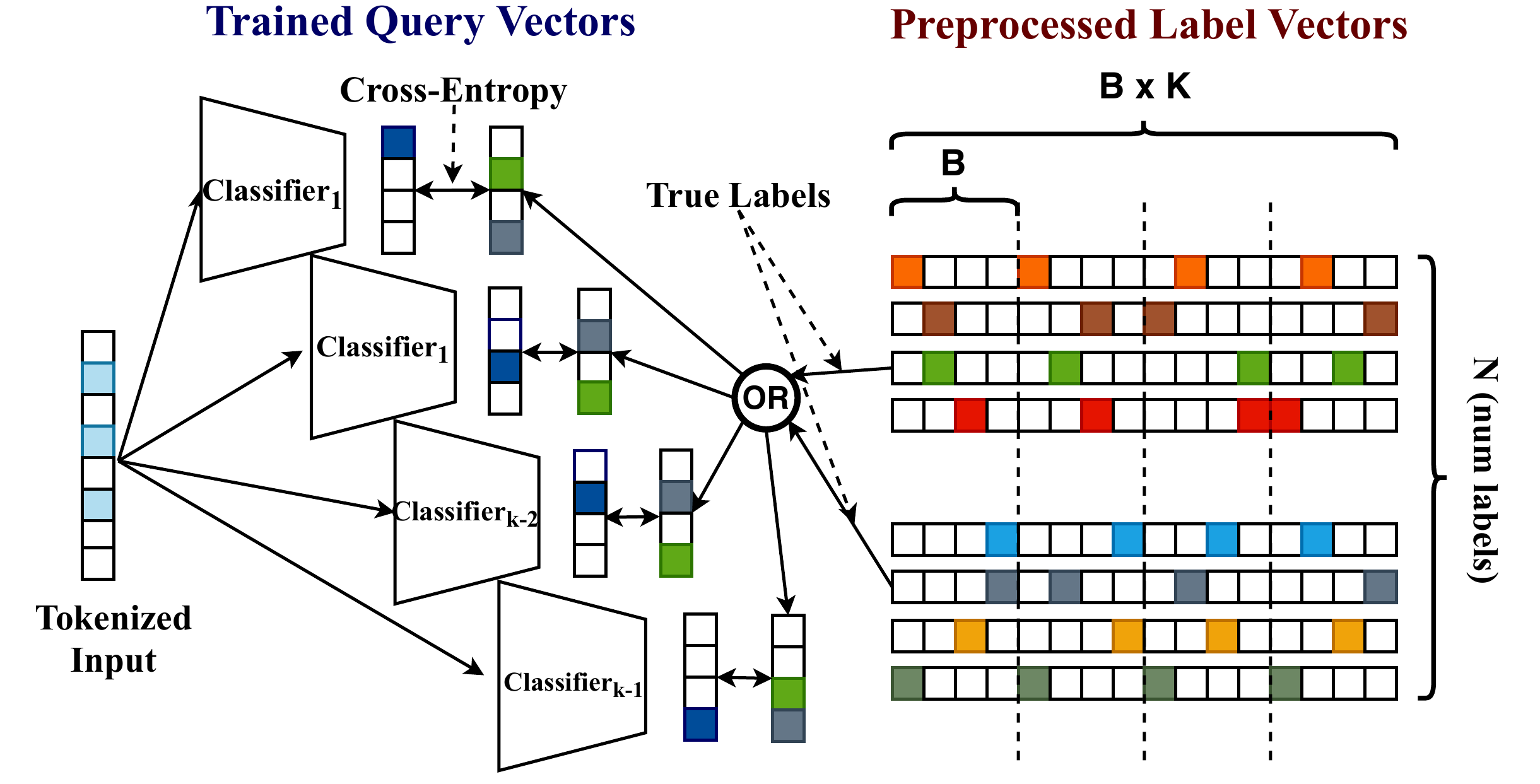}
    \caption{Schematic diagram for label vector construction (on the right) and the training process (on the left). Each label vector is $B \times K$ dimensional divided into $K$ components of length $B$. Each vector is $K$-sparse with exactly one non-zero index in each component (colored on the right). The components are separated by dotted vertical lines. For a given input, we perform an `OR' operation over the true label vectors and feed the resultant pieces to independent small classifiers.}
    \label{fig:training}
    \vspace{-0.3cm}
\end{figure}

\vspace{-0.3cm}
\section{Our Method: SOLAR} \label{sec:SOLAR}
\vspace{-0.2cm}
In this section, we describe in detail the workflow of our algorithm SOLAR. First, we will discuss the pre-processing phase where we construct random sparse label vectors (figure \ref{fig:training}) and an inverted-index of the labels (figure \ref{fig:inv_idx}). Then, we move to the training phase where we split the label vectors into independent contiguous components and train each of them in parallel (figure \ref{fig:training}). In the end, we show the inference procedure where we obtain the piece-wise query vector in parallel and sparsify by retaining only top buckets from each piece. We then look up the saved inverted index to retrieve and score the candidate labels to sort and predict the best ones (figure \ref{fig:inference}).

{\bf Notations:} $N$ denotes the total number of labels. $D$ is the sparse vector dimension. $K$ is the number of non-zeros in label vectors. $B=\frac{D}{K}$ is the number of buckets in each component of the vector.

\subsection{Pre-processing: Construction of Label Embeddings and Inverted-Index}\label{subsec:construction}
As presented in figure~\ref{fig:training}, let there be $N$ labels ($N$ is large, in the order of a million). We intend to construct $K$-sparse (having $K$ non-zero indices) high dimensional vectors for each label. As noted earlier, a large output dimension makes training a cross-entropy loss prohibitively expensive. Therefore, inspired by recent work on zero-communication Model Parallelism~\citep{mach}, we partition the large dimensional vector into $K$ subsets and train each one independently. Each subset of the partition comprises of $B$ buckets with exactly one non-zero index. The colored blocks on the right side in figure \ref{fig:training} denote the non-zero indices for each label vector.
 
To adhere to our design principle of load-balancing, for each label, we pick the non-zero index randomly in the range of $B$ for each of the $K$ components. To be precise, for any label, we randomly generate $K$ integers in the range of $B$. As in most of our experiments, set $K=16$ and $B=30K$. This makes the overall dimension $D = B \times K = 480K$ and a sparsity ratio of $0.000533$ ($0.0533\%$). As an example, let the generated integers be $\{18189,8475,23984,....,17924,459\}$. Then the non-zero indices of the overall vector are simply $B$-shifted , i.e., $\{18189,38475,83984,....,437924,450459\}$. Although any random number generator would work fine, we pick our non-zero indices using {\it sklearn}'s {\it murmurhash} function. It is rather straightforward to see that these vectors are near-orthogonal. The expected dot-product between any two label vectors $l_i$ and $l_j$ is, 
\begin{equation}
 E({l_i}^T * l_j)\ =\ \sum_k p(h_k(i)=h_k(j))\ =\ \frac{K}{B}\approx\ 0   .
\end{equation}

\begin{wrapfigure}{}{6.01cm}
    \vspace{-0.6cm}
    \centering
    \includegraphics[width=5.0cm]{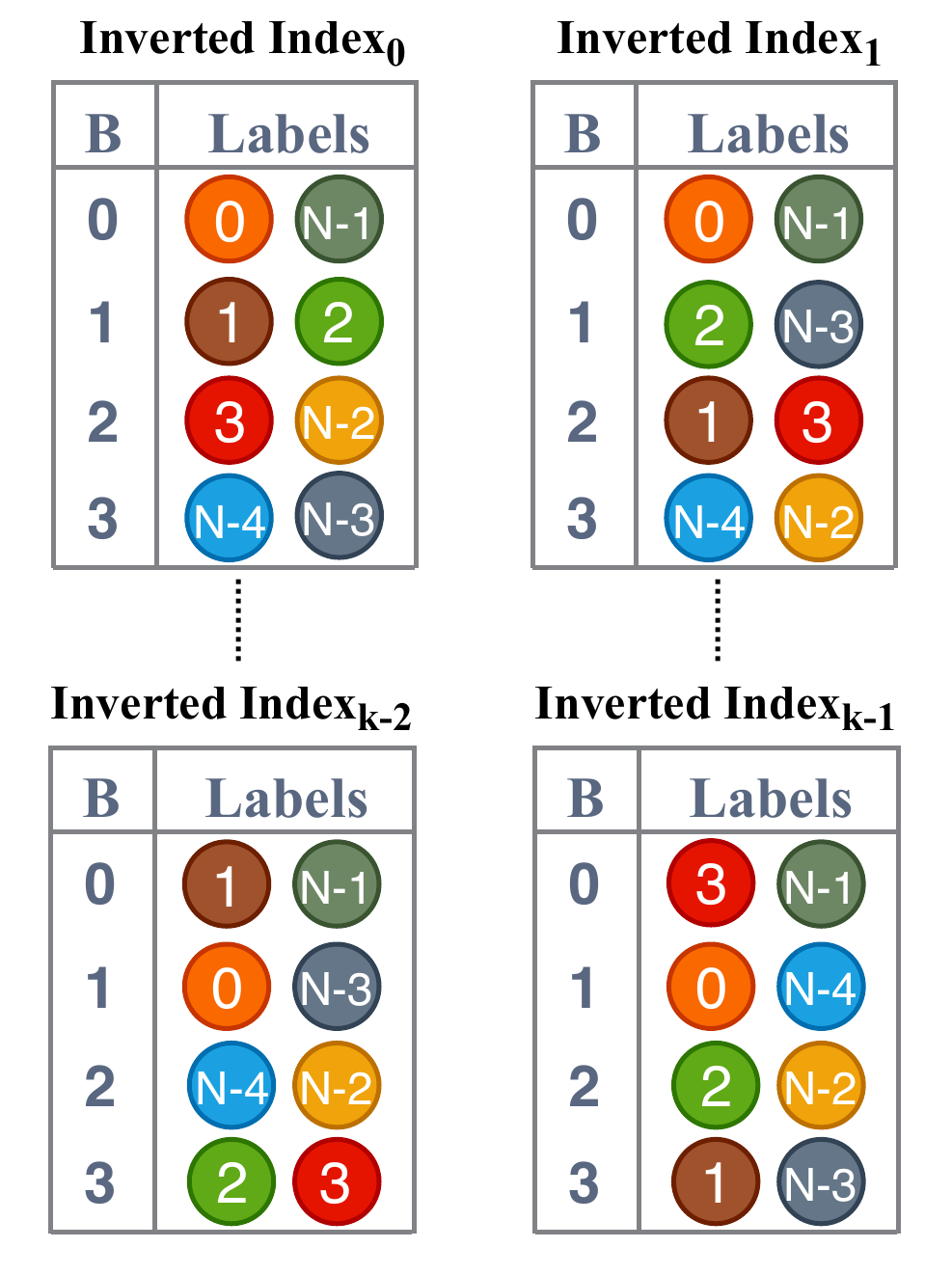} %
    \caption{Inverted-Index construction for the label vectors shown in figure \ref{fig:training}. We construct one index for each of the $K$ chunks. Each bucket will have the same number of labels by design (Load-Balanced).}
    \label{fig:inv_idx}
    \vspace{-0.2cm}
\end{wrapfigure}

Figure \ref{fig:inv_idx} shows the toy inverted index for the label vectors shown in figure \ref{fig:training}. Since we train $K$ independent models, each model is expected to predict its own `buckets of high relevance'. Hence we maintain $K$ separate inverted-indexes. For any input, we accumulate the candidates from each of the $K$ inverted-indexes and take a union of them for scoring and sorting. It is noteworthy that two unrelated labels might be pooled into the same bucket. While this sounds rather jarring from a learnability perspective, it is essential for the load-balance and also to learn a positive-only association of input tokens and true-label buckets (more on this Appendix B).

\vspace{-0.3cm}
\subsection{Training}\label{subsec:training}
\vspace{-0.2cm}

Figure \ref{fig:training} also depicts the training process (on the left side). In a multilabel learning problem, each input has a variable number of true labels. We lookup all the true label vectors for an input and perform an `OR' operation over the respective sparse label vectors. Please note that at the level of sparsity we are dealing, even with zero pairwise collisions among the non-zero indices of label vectors, we still have a super-sparse representation for the resultant `OR' vector. We partition this combined-label vector into $K$ parts just like before and train individual classifiers (simple feed forward neural networks with 1 hidden layer) with a binary cross entropy loss function with the $B$ dimensional few-hot vectors. Please note that these models do not communicate with each other. Since there is no overhead of parameter sharing, training can be embarrassing parallellized across multiple GPUs (Zero-Communication Model Parallellism).

\indent {\bf Input Feature Hashing:} Usually, naive input tokenization like bag-of-words (BoW) leads to a very high dimensional input. This in turn makes the first layer of the network intractable. Hence, an elegant solution for this problem is to hash the token indices to a lower dimension (called Feature Hashing~\citep{feathash}). In our case, we use a different random seed for each of the $K$ models and hash the input indices to a feasible range. Although we lose some input-information in each individual model (due to feature hash collisions), the variation in random seed minimizes this loss when all the models are collectively taken into account.

\vspace{-0.3cm}
\subsection{Inference}\label{subsec:inference}
\vspace{-0.2cm}

\begin{wrapfigure}{}{9.05cm}
\vspace{-1.0cm}
    \centering
    \includegraphics[width=9.0cm]{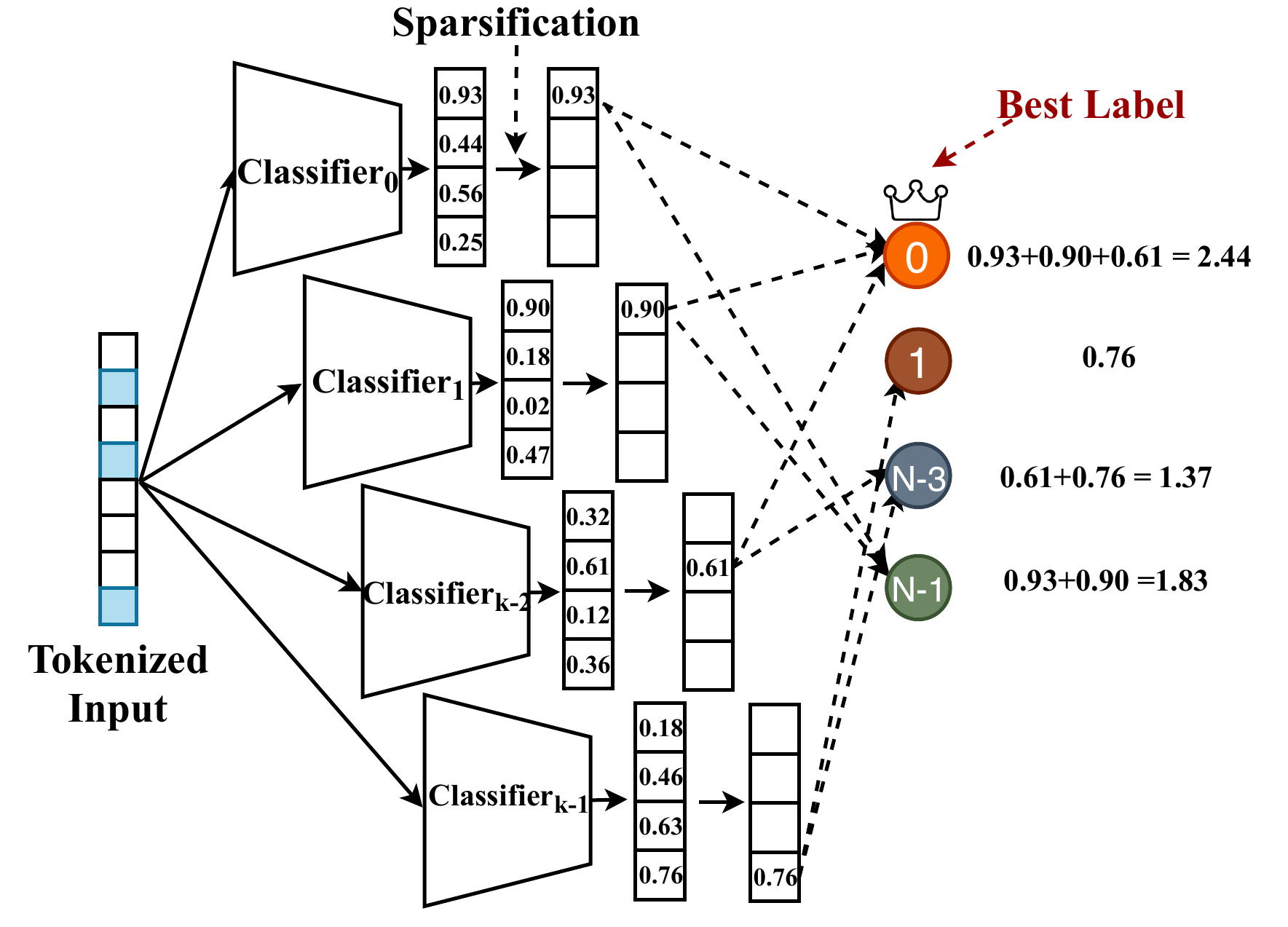} %
    \caption{Schematic diagram for Inference. We first get $K$ probability vectors of $B$ dimensions each. Then we only retain the top-m buckets after sparsification (m=1 in above figure. For our experiments, m varies among 50 and 100). We accumulate the candidate labels based on inverted-index for these top-buckets and aggregate their scores and identify the best labels.}
    \label{fig:inference}
    \vspace{-0.4cm}
\end{wrapfigure}

One of the key advantages of SOLAR over dense embedding models is faster inference. As mentioned earlier, the primary reason for this is the replacement of matrix-multiplication and sorting with simple lookups and aggregations. This workflow is depicted in figure \ref{fig:inference}. Given a tokenized input, we pass it through all the $K$ models in parallel and obtain the respective $B$ dimensional probability vectors. We then sort these probabilities and obtain the top-$m$ ($m$ varies among 50 and 100) buckets for each model. These $m \times K$ integers constitute the non-zero indices of the SOLAR embedding for the input query. We can query the $K$ inverted-indexes with the respective top-$m$ buckets for candidates. A union of all these candidates is our target set of labels. For each of these candidates, we sum up the predicted probability scores from the corresponding buckets and sort for the top results.

\textbf{Time-Complexity:} Since our inverted indexes are load-balanced, each bucket accommodates $\frac{N}{B}$ labels. Hence, the top-$m$ buckets contribute $\frac{mN}{B}$ candidates. 
The candidates retrieved from $K$ models are bound to have some overlapping labels, accompanied by some unrelated counterparts from the respective buckets (since we randomly initialized the label vectors). 

In the worst case of zero overlaps, the total number of candidates would be $\frac{KmN}{B}$. The aggregation of scores is a by-product of the candidate selection step. Finally, we sort the aggregated scores for these candidates. Including the two sorting steps: 1) to get top-$m$ buckets 2) to get top $5$ labels, the total number of operations performed is $B\log m + \frac{KmN}{B} + \frac{KmN}{B}\log 5$. 

A dense embedding model on the other hand needs $NmK + N \log 5$ steps (assuming dense vectors of dimension $d\ =\ mK$, since SOLAR also has $mK$ non-zeros after inference). For the scale of $N$ and $mK$ we are dealing with ($N=1M$, $K=16$, $m=50$, $B=30K$), SOLAR supposedly has $\frac{B}{1+\log 5}$ times faster inference. However, matrix multiplications have specialized processes with hardware acceleration unlike SOLAR and the practical gains would be in the order of 5x (more in section \ref{subsec:p2p}).
\vspace{-0.3cm}
\section{Analysis: Is one-sided learning all we need?}\label{sec:onesided}
\vspace{-0.3cm}
As argued before, we need distributed training to learn ultra-high dimensional embeddings. Our distributed process requires randomly initializing and fixing label embeddings, which might seem to be a questionable approach. However, it turns out that this method of learning is mathematically equivalent to the standard two-sided learning approach with orthogonality constraints. Enforcing such orthogonality in the embedding space for information gain is a common practice~\citep{zhang2017learning, glas}. 
More specifically, we prove that the following two processes are mathematically equivalent under any kernel: 1) learning standard two-sided embedding (for both inputs and labels) with orthogonality constraint on the label embedding 2) starting with any fixed orthogonal embedding in the label space and learning the input embeddings. This is a nuanced analogy to the `Principle of Deferred Decision'.

An embedding model comprises of two functions $f_I$ and $f_L$ which map the inputs and labels respectively to a common vector space. Let $X$ be the set of all inputs and $Y$ be the set of all labels. For some $x\in X$ and $y\in Y$, we seek to optimize $f_I$ and $f_L$ such that the inner product $\langle f_I(x),f_L(y) \rangle$ (or any kernel) is the desired similarity metric $S(x,y)$. Typically, $f_I$ and $f_L$ are neural networks that map tokenized sentences to a vector space and $S(x,y)\ =\ \langle f_I(x),f_L(y) \rangle\ =\ \frac{f_I(x)^Tf_L(y)}{\Vert f_I(x)\Vert_2 \Vert f_L(y)\Vert_2}$ (cosine similarity). A special case of the embedding models are the popular Siamese Networks~\citep{siamese, dssm, dlrm} in which $f_I=f_L$, i.e., both inputs and outputs learn a shared embedding space.

The imposition of orthogonality on the label vectors supposedly learns a function $f_L$ such that
$$\langle f_L(y_i),f_L(y_j)\rangle = \delta_{ij}\ \forall y_i,y_j\in Y$$
where $\delta_{ij}$ is the Kronecker-delta function. Hence, $\{f_L(y_1),f_L(y_2),f_L(y_3),...\}$ form an orthonormal basis (since we ensure that the learned vectors are unit norm). The following theorem states that both the aforementioned learning methods are equivalent.

\begin{theorem} {\bf Law of Deferred Orthogonality of Learned Embedding}
Given any positive semi-definite kernel $S(.,.)$, and functions $f_I$ and $f_L$, where $f_L$ is orthogonal. For any orthogonal function $R$ with the same input and range space as $f_L$, there always exist a function $f$, such that
$S(f_I(x),f_L(y)) = S(f(x),R(y)).$
\label{thm:orthogonal}
\end{theorem}
\begin{proof}
\vspace{-0.2cm}
Please refer to Appendix C for detailed proof. The result mainly follows from lemma~\ref{lemma:1}.
\vspace{-0.1in}
\end{proof}
 
\begin{lemma}\label{lemma:1}
For any two orthonormal basis matrices, $\bf A = [\overline{\bf a_1}\ \overline{\bf a_2}\ \overline{\bf a_3}\ ...\ \overline{\bf a_n}]$ and $\bf B = [\overline{\bf b_1}\ \overline{\bf b_2}\ \overline{\bf b_3}\ ...\ \overline{\bf b_n}]$ in the same vector space, there exists an orthogonal matrix $\bf P$ such that $\bf A = \bf P\bf B$.
\end{lemma}
Note that, since we construct fixed norm label vectors, orthogonal vectors form orthonormal basis vectors (up to a constant norm factor multiplication).

{\bf Power of Randomness:} The popular law of deferred decision in probability allows us to pre-generate random numbers for efficiency. In an analogous fashion, Theorem~\ref{thm:orthogonal} allows us to pick any orthogonal transformation $R$ in advance and only learn one-sided embedding function $f$. By exploiting this flexibility we design random, binary, and ultra-sparse label vectors which make training affordable at $500K$ dimensions. 

\vspace{-0.3cm}
\section{Experiments}\label{sec:experiments}
\vspace{-0.2cm}
We now validate our method on two main tasks 1) Product to Product Recommendation on a 1.67M book dataset. This dataset simulates the typical product recommendation task in most search engines. 2) Extreme Classification with the three largest public datasets. Most of the deployed models on modern search engines show promising results on these public datasets in addition to their respective private datasets~\citep{parabel,slice,mach}. 
\vspace{-0.3cm}
\subsection{Product to Product Recommendation}\label{subsec:p2p}
\vspace{-0.2cm}
{\bf Dataset:} This dataset is curated from the raw Amazon Book dataset on extreme classification repository (XML-Repo)~\citep{xml_repo}. This dataset comprises of 1604777 training books whose titles serve as input queries. Each query book is mapped to a set of target books serving as labels (related products recommendation problem). There are a total of 1675657 label books with titles. After parsing both the query titles and label titles, the total vocabulary comprises of 763265 words. The average query length is $\sim$ 9 for both input titles and label titles. There are additionally 693K eval books.

{\bf Hyperparameters:} As mentioned before, we train 480K dimensional SOLAR embeddings split into $K=16$ chunks of $B=30K$ buckets each. The label embeddings are fixed to be exactly 16-sparse while the learned query embeddings are evaluated with $1600$ non-zero indices (by choosing $m=100$ top buckets). We feature hash the 763265-dimensional BOW inputs to 100K dimensions. Each independent model is a feed-forward network with an input layer with 100K nodes, one hidden layer with 4096 nodes, and an output layer with $B=30K$ nodes. For minimizing the information loss due to feature hashing, we choose a different random seed for each model. Note that these random seeds have to be saved for consistency during evaluation.

{\bf Machine and Frameworks:} We train with Tensorflow (TF) v1.14 on an DGX machine with 8 NVIDIA-V100 GPUS. We use TF Records data streaming to reduce GPU idle time. The accompanying code has detailed instructions for all the 3 steps. During training, we use a batch size of 1000. During inference, except getting the sparsified probability scores, all other steps are performed on CPU using python's $multiprocessing$ module with 48 threads. Our metrics of interest are standard precision and recall. We measure precision at 1,5,10 and recall at 100 (denoted by P@1, P@5,P@10 and Rec@100).

{\bf Baselines:} The most natural comparison arises with recent industry standard embedding models, namely Amazon's Deep Semantic Search Model (DSSM)~\citep{dssm} and Facebook's Deep Learned Recommendation Model (DLRM)~\citep{dlrm}. DLRM takes mixed inputs (dense and sparse) and learns token embeddings through a binary classifier for ad prediction. On the other hand, DSSM trains embeddings for semantic matching where we need to suggest relevant products for a user typed query. We compare against DSSM as it aligns with this dataset. Additionally, we impose orthogonality in DSSM using GlaS~\citep{glas} regularizer. We also compare against other embedding models AnnexML~\citep{annexml} and SNRM~\citep{snrm}. Please refer to Appendix D for baseline settings.

{\bf Results:} Table \ref{tab:result_p2p} compares SOLAR against all the aforementioned baselines. The top row corresponds to the evaluation of SOLAR without any sparsification (hence called SOLAR-full). The second row corresponds to the case where we pick the top-100 buckets in each of the 16 models (and hence an 1600 sparse vector and a sparsity ratio of 1600/480K = 0.333\%). We notice that on all the metrics, SOLAR is noticeably better than DSSM and its improved variant with a GLaS regularizer. SNRM clearly underperforms due to reasons mentioned in section \ref{sec:rel_work}. An interesting point to note is that the evaluation for DSSM was totally done on GPU while SOLAR spends most of its time on CPU. Despite that, SOLAR infers much faster than other baselines.

Please refer to Appendix D for some sample evaluation queries which show that DSSM works well for frequently queried books while it falters for the tail queries. SOLAR on the other hand is more robust to this phenomenon. Additionally, we also examine the performance of SOLAR and DSSM with new unseen products (an analogue to zero-shot Learning).

\begin{table*}[t]
\resizebox{\linewidth}{!}{
\renewcommand{\arraystretch}{1.15}
\begin{tabular}{ | c | c | c | c | c | c | c |c|}
\hline
Model & epochs & P@1 & P@5 & P@10 & Rec@100 & Train Time (hrs) & Eval Time (ms/point)\\
\hline
SOLAR (full) & 10 & 43.12 & 40.44 & 40.02  & 50.48 & 2.65 & 5.09\\
\hline
SOLAR (m=100) & 10 & {\bf 35.24} & {\bf 29.71} & {\bf 26.98}  & {\bf 34.18} & {\bf 2.65} & {\bf 0.96}\\
\hhline{|=|=|=|=|=|=|=|=|}
DSSM (d=1600) & 5 & 31.34 & 27.55 & 24.41  & 32.71 & 25.27 & 1.77 \\
\hline
GLaS (d=1600) & 5 & 32.51 & 28.31 & 25.41 & 33.17 & 37.14 & 1.77 \\
\hhline{|=|=|=|=|=|=|=|=|}
SNRM (d=30K) & 5 & 1.59 & 2.01 & 1.93 & 2.41 & - & -\\
\hhline{|=|=|=|=|=|=|=|=|}
AnnexML (d=800) & 10 & 26.31 & 22.22 & 19.37 & 26.13 & 16 & 3.06\\
\hline
\end{tabular}
}
\caption{Comparison of SOLAR against DSSM, DSSM+GLaS, and SNRM baselines. SOLAR (full) corresponds to evaluating all labels without shortlisting candidates. Only the best among the sparse models is highlighted. SOLAR's metrics are better than the industry-standard DSSM model while training 10x faster and evaluating 2x faster (SOLAR-CPU vs DSSM-GPU evaluation). GLaS regularizer improves the metrics but still lags behind SOLAR.}
\label{tab:result_p2p}
\end{table*}

\begin{table*}[t]
\resizebox{\linewidth}{!}{
\renewcommand{\arraystretch}{1.15}
\begin{tabular}{|c|c|c|c|c|c|c|c|c|c|}
\hline
\multicolumn{2}{|c|}{}& \multicolumn{5}{c|}{Embedding Models} & \multicolumn{3}{c|}{Other Baselines} \\
\hline
Dataset & Metric & SOLAR (full) & SOLAR (m=100) & SOLAR (m=50) & AnnexML & SLEEC & Parabel & Pfastre XML & SLICE\\
\hhline{|=|=|=|=|=|=|=|=|=|=|}
\multirow{3}{*}{Wiki-500K} & P@1 & 62.43 & {\bf 60.92} & 60.52 & 56.81 & 30.86 & 59.34 & 55.00 & 59.89\\
\cline{2-10}
 & P@3 & 48.16 & {\bf 46.94} & 46.56 & 36.78 & 20.77 & 39.05 & 36.14 & 39.89\\
\cline{2-10}
 & P@5 & 45.40 & {\bf 45.32} & 45.28 & 27.45 & 15.23 & 29.35 & 27.38 & 30.12\\
\hhline{|=|=|=|=|=|=|=|=|=|=|}

\multirow{3}{*}{Amz-670K} & P@1 & 37.66 & {\bf 34.37} & 34.19 & 26.36 & 18.77 & 33.93 & 28.51 & \underline{37.77}\\
\cline{2-10}
 & P@3 & 35.30 & {\bf 32.71} & 32.51 & 22.94 & 16.5 & 30.38 & 26.06 & \underline{33.76}\\
\cline{2-10}
 & P@5 & 33.99 & {\bf 32.55} & 32.46 & 20.59 & 14.97 & 27.49 & 24.17 & 30.70\\
\hhline{|=|=|=|=|=|=|=|=|=|=|}

\multirow{3}{*}{Amz-3M} & P@1 & 47.70 & {\bf 44.89} & 44.61 & 41.79 & -  & \underline{47.51} & 43.83 & -\\
\cline{2-10}
 & P@3 & 44.58 & {\bf 42.36} & 42.08 & 38.24 & - & \underline{44.68} & 41.81 & -\\
\cline{2-10}
 & P@5 & 43.53 & {\bf 41.03} & 40.69 & 35.98 & - & \underline{42.58} & 40.09 & -\\
\hline
\end{tabular}
}
\caption{SOLAR vs popular Extreme Classification benchmarks. Embedding models AnnexML and SLEEC clearly underperform compared to SOLAR. SOLAR even outperforms the state-of-the-art non-embedding baselines like Parabel and Slice. The gains in P@5 are particularly huge (45.32\% vs 31.57\%). SLEEC and SLICE do not scale up to 3M labels (corroborated on XML-Repo).}
\label{tab:result_xml}
\end{table*}



\begin{table*}[t]
\resizebox{\linewidth}{!}{
\renewcommand{\arraystretch}{1.15}
\begin{tabular}{|c|c|c|c|c|c|c|}
\hline
Dataset &  & SOLAR (m=100) & SOLAR (m=50) & SLICE & Parabel & Pfastre XML\\
\hhline{|=|=|=|=|=|=|=|}
\multirow{2}{*}{Wiki-500K} & Training (hrs) & 2.52  & 2.52 & {\bf 2.34} & 6.29 & 11.14 \\
\cline{2-7}
 & Eval (ms/point) & 1.1 & {\bf 0.76} & 1.37 & 2.94 & 6.36 \\
\hhline{|=|=|=|=|=|=|=|}
\multirow{2}{*}{Amz-670K} & Training (hrs) & {\bf 1.19} & {\bf 1.19} & 1.92 & 1.84 & 2.85 \\
\cline{2-7}
 & Eval (ms/point) & 2.56 & {\bf 1.58} & 3.49 & 2.85 & 19.35\\
\hhline{|=|=|=|=|=|=|=|}

\multirow{2}{*}{Amz-3M} & Training (hrs) & 5.73 & 5.73 & - & {\bf 5.39} & 15.74\\
\cline{2-7}
 & Eval (ms/point) & 2.09 & 1.87 & - & {\bf 1.72} & 4.05\\
\hline
\end{tabular}
}
\caption{Training and Evaluation speeds against the fastest baselines.}
\label{tab:time_xml}
\end{table*}

\begin{table*}[h]
\resizebox{\linewidth}{!}{
\renewcommand{\arraystretch}{1.15}
\begin{tabular}{|c|cccc|cccc|cccc|}
\hline
 & \multicolumn{4}{c|}{K=8} & \multicolumn{4}{c|}{K=16} & \multicolumn{4}{c|}{K=32}\\
\cline{2-13} &P@1&P@3 & P@5 & Time (ms)  & P@1&P@3 & P@5 & Time (ms)  & P@1&P@3 & P@5 & Time (ms)  \\
\hline
B=30K & 31.34&29.23&28.45 & 0.647  & 33.27&31.34&30.90 & 0.908   & 33.99&32.39&32.31& 1.65  \\
\hline
B=20K & 30.2&28.25&27.31& 0.621  & 32.55&30.69&29.99 & 0.96 &  {\bf 33.74} & {\bf 32.06} & {\bf 31.78 }& {\bf 1.52 }\\
\hline 
B=10K & 27.99&25.58&24.36 &0.97 & 29.72&27.85&27.04 &1.326  & 32.07&30.22&29.65 &1.39 \\
\hline
\end{tabular}
}
\caption{Effect of $B,K$ on $P@1/3/5$ for Amz-670K dataset (with $m=25$). We increment $B$ linearly and $K$ exponentially and choose an optimal trade-off between precision and inference time (shown in ms/point).}
\label{tab:ablation}
\vspace{-0.3cm}
\end{table*}

\vspace{-0.3cm}
\subsection{Extreme Classification Datasets}\label{subsec:xml}
\vspace{-0.2cm}
Let us now shift our focus to the 3 largest multi-label learning datasets available on the XML-Repo, namely Amazon-3M, Amazon-670K and Wiki-500K datasets with 3M, 670K and 500K labels respectively. The statistics of these datasets are available on the repository. 

{\bf Hyper-parameters:} For Amazon-3M, the hyper-parameters remain the same as the book recommendation dataset. For the Wiki-500K and Amazon-670K datasets, we use the latest versions with dense 512-dimensional inputs as outlined in~\citep{slice}. Since the input is now much lower-dimensional than the sparse versions of the same datasets, we train $K=32$ models in parallel with $B=20K$, thereby making the overall dimension 640K as opposed to the 480K before. We report the standard P@1, P@3, and P@5 metrics and perform two levels of sparsification for all 3 datasets; $m=50$ in addition to the previous $m=100$.  

{\bf Baselines:} Since the labels do not have annotations for these datasets, we cannot aspire to train Siamese models like DSSM and SNRM here. Hence we choose to compare against the popular embedding models AnnexML~\citep{annexml} and SLEEC~\citep{sleec} in addition to other Extreme Classification benchmarks like the pure tree-based PfastreXML~\citep{Pfastre}, tree and 1-vs-all model Parabel~\citep{parabel} and also against the recent NN-graph based SLICE~\citep{slice} which is the state-of-the-art for the first 2 datasets.

{\bf Results:} Comparison of precision for SOLAR against both embedding and non-embedding baselines are shown in table \ref{tab:result_xml}. We use the reported numbers for all available baselines. However, AnnexML and SLEEC do not have reported scores for the 2 smaller datasets with dense inputs. We run the official C++ and MATLAB packages for either of them. It is noteworthy that the training and evaluation for SLEEC are rather slow even though the model size is smaller with dense inputs. Hence, SLEEC could not be scaled to the large 3M class dataset. This fact is independently verified on the repository. It is clear that SOLAR outperforms the state-of-the-art baselines, including SLICE which is noteworthy because SLICE has been very successful on Bing Search for query suggestion (improving the trigger coverage by 52\%, mores so for tail queries).

The speed comparison for SOLAR against the fastest baselines is shown in table \ref{tab:time_xml}. We can see that SOLAR either matches or surpasses the best ones on both training and inference. Parabel closes in on SOLAR on Amz-3M while SOLAR is much faster on Wiki-500K. PfastreXML is slower than the rest.

{\bf Ablation Study on Choice of $B$ and $K$:} Choosing an appropriate number of partitions $K$ and buckets $B$ is very important as it causes a trade-off between precision and speed. We usually have diminishing gains in precision with increasing $B$ and $K$ at a cost of training and inference speeds. Hence, to strike a good balance, we start with $B=10K$ buckets and increment by $10K$ until the gains are insignificant. Similarly, we increase $K$ in powers of 2 until convergence in precision performance. Table \ref{tab:ablation} shows the performance trend for Amazon-670K dataset. We notice that $B=20K,\ K=32$ is an optimal setting for this dataset.

{\bf Memory Advantage:} Dense embedding models need to hold all label embeddings in GPU memory to perform real-time inference. This is not a scalable solution when the labels run into the order of 100 million (which is a practical industry requirement). With an embedding dimension of 1000, we would need to 100 billion fp32 parameters which take up 800 GB of GPU memory. Hence, most practical recommender systems store label vectors on CPU which would make real-time inference infeasible. On the contrary, SOLAR needs to store only 16 integers per label which is very memory efficient with modern sparse array support on all platforms.
 
\section{Conclusion}
This paper proposes a first-of-its-kind algorithm SOLAR to learn high-dimensional sparse vectors against dense low-dimensional ones. Sparse vectors are conducive to efficient data structures like inverted-indexes for large scale Information Retrieval. However, training high dimensional vectors has major bottlenecks. Through some elegant design choices, SOLAR ensures that the high dimensional embeddings are trainable in a distributed fashion with Zero-Communication. Additionally, SOLAR enforces near equal inference times for all queries by load-balancing the inverted-indexes. When applied to a multitude of product recommendations and extreme classification datasets, SOLAR outperforms the respective state-of-the-art methods by a large margin on precision and speed.

\vspace{-0.2cm}
\section*{Broader Impact}
\vspace{-0.2cm}
This paper proposes a new paradigm of embedding models for very Large Scale Information Retrieval (IR) applications. About 75\% of Deep Learning models trained on cloud are for applications like recommending products, advertisements, food, apparel, \emph{etc}. Our method is one direction towards reducing the training and inference costs for several of these applications, manifesting in huge savings in energy bills for major cloud AI platforms. 

Further, the world of NLP has been stuck to dense embedding models for a very long time now. This work can potentially be used to train sparse word and sentence vectors and challenge state-of-the-art models in Language Translation. SOLAR learns a positive-only correlation between tokens and that can be mighty useful to train a One-Model-for-all-Languages kind of translator which can take multiple input language queries and translate to a single target language.

\bibliography{main}

\begin{thebibliography}{33}
\providecommand{\natexlab}[1]{#1}
\providecommand{\url}[1]{\texttt{#1}}
\expandafter\ifx\csname urlstyle\endcsname\relax
  \providecommand{\doi}[1]{doi: #1}\else
  \providecommand{\doi}{doi: \begingroup \urlstyle{rm}\Url}\fi

\bibitem[Croft et~al.()Croft, Metzler, and Strohman]{croft2010search}
W~Bruce Croft, Donald Metzler, and Trevor Strohman.
\newblock \emph{Search engines: Information retrieval in practice}, volume 520.

\bibitem[Baeza-Yates et~al.(1999)Baeza-Yates, Ribeiro-Neto,
  et~al.]{baeza1999modern}
Ricardo Baeza-Yates, Berthier Ribeiro-Neto, et~al.
\newblock \emph{Modern information retrieval}, volume 463.
\newblock ACM press New York, 1999.

\bibitem[Mikolov et~al.(2013)Mikolov, Sutskever, Chen, Corrado, and
  Dean]{word2vec}
Tomas Mikolov, Ilya Sutskever, Kai Chen, Greg~S Corrado, and Jeff Dean.
\newblock Distributed representations of words and phrases and their
  compositionality.
\newblock In \emph{Advances in neural information processing systems}, pages
  3111--3119, 2013.

\bibitem[JeffreyPennington and Manning()]{glove}
RichardSocher JeffreyPennington and ChristopherD Manning.
\newblock Glove: Global vectors for word representation.
\newblock Citeseer.

\bibitem[Vaswani et~al.(2017)Vaswani, Shazeer, Parmar, Uszkoreit, Jones, Gomez,
  Kaiser, and Polosukhin]{attention}
Ashish Vaswani, Noam Shazeer, Niki Parmar, Jakob Uszkoreit, Llion Jones,
  Aidan~N Gomez, {\L}ukasz Kaiser, and Illia Polosukhin.
\newblock Attention is all you need.
\newblock In \emph{Advances in neural information processing systems}, pages
  5998--6008, 2017.

\bibitem[Devlin et~al.(2018)Devlin, Chang, Lee, and Toutanova]{bert}
Jacob Devlin, Ming-Wei Chang, Kenton Lee, and Kristina Toutanova.
\newblock Bert: Pre-training of deep bidirectional transformers for language
  understanding.
\newblock \emph{arXiv preprint:1810.04805}, 2018.

\bibitem[Nigam et~al.(2019)Nigam, Song, Mohan, Lakshman, Ding, Shingavi, Teo,
  Gu, and Yin]{dssm}
Priyanka Nigam, Yiwei Song, Vijai Mohan, Vihan Lakshman, Weitian Ding, Ankit
  Shingavi, Choon~Hui Teo, Hao Gu, and Bing Yin.
\newblock Semantic product search.
\newblock In \emph{Proceedings of the 25th ACM SIGKDD International Conference
  on Knowledge Discovery \& Data Mining}, pages 2876--2885, 2019.

\bibitem[Naumov et~al.(2019)Naumov, Mudigere, Shi, Huang, Sundaraman, Park,
  Wang, Gupta, Wu, Azzolini, et~al.]{dlrm}
Maxim Naumov, Dheevatsa Mudigere, Hao-Jun~Michael Shi, Jianyu Huang, Narayanan
  Sundaraman, Jongsoo Park, Xiaodong Wang, Udit Gupta, Carole-Jean Wu,
  Alisson~G Azzolini, et~al.
\newblock Deep learning recommendation model for personalization and
  recommendation systems.
\newblock \emph{arXiv preprint arXiv:1906.00091}, 2019.

\bibitem[Prabhu and Varma(2014)]{prabhu2014fastxml}
Yashoteja Prabhu and Manik Varma.
\newblock Fastxml: A fast, accurate and stable tree-classifier for extreme
  multi-label learning.
\newblock In \emph{Proceedings of the 20th ACM SIGKDD international conference
  on Knowledge discovery and data mining}, pages 263--272, 2014.

\bibitem[Jain et~al.(2016)Jain, Prabhu, and Varma]{Pfastre}
Himanshu Jain, Yashoteja Prabhu, and Manik Varma.
\newblock Extreme multi-label loss functions for recommendation, tagging,
  ranking \& other missing label applications.
\newblock In \emph{Proceedings of the 22nd ACM SIGKDD International Conference
  on Knowledge Discovery and Data Mining}, pages 935--944, 2016.

\bibitem[Agrawal et~al.(2013)Agrawal, Gupta, Prabhu, and
  Varma]{agrawal2013multi}
Rahul Agrawal, Archit Gupta, Yashoteja Prabhu, and Manik Varma.
\newblock Multi-label learning with millions of labels: Recommending advertiser
  bid phrases for web pages.
\newblock In \emph{Proceedings of the 22nd international conference on World
  Wide Web}, pages 13--24, 2013.

\bibitem[Weston et~al.(2013)Weston, Makadia, and Yee]{weston2013label}
Jason Weston, Ameesh Makadia, and Hector Yee.
\newblock Label partitioning for sublinear ranking.
\newblock In \emph{International conference on machine learning}, pages
  181--189, 2013.

\bibitem[Yen et~al.(2016)Yen, Huang, Ravikumar, Zhong, and Dhillon]{pdsparse}
Ian En-Hsu Yen, Xiangru Huang, Pradeep Ravikumar, Kai Zhong, and Inderjit
  Dhillon.
\newblock Pd-sparse: A primal and dual sparse approach to extreme multiclass
  and multilabel classification.
\newblock In \emph{International Conference on Machine Learning}, pages
  3069--3077, 2016.

\bibitem[Yen et~al.(2017)Yen, Huang, Dai, Ravikumar, Dhillon, and
  Xing]{ppdsparse}
Ian~EH Yen, Xiangru Huang, Wei Dai, Pradeep Ravikumar, Inderjit Dhillon, and
  Eric Xing.
\newblock Ppdsparse: A parallel primal-dual sparse method for extreme
  classification.
\newblock In \emph{Proceedings of the 23rd ACM SIGKDD International Conference
  on Knowledge Discovery and Data Mining}, pages 545--553, 2017.

\bibitem[Bi and Kwok(2013)]{bi2013efficient}
Wei Bi and James Kwok.
\newblock Efficient multi-label classification with many labels.
\newblock In \emph{International Conference on Machine Learning}, pages
  405--413, 2013.

\bibitem[Tagami(2017)]{annexml}
Yukihiro Tagami.
\newblock Annexml: Approximate nearest neighbor search for extreme multi-label
  classification.
\newblock In \emph{Proceedings of the 23rd ACM SIGKDD international conference
  on knowledge discovery and data mining}, pages 455--464, 2017.

\bibitem[Bhatia et~al.(2015)Bhatia, Jain, Kar, Varma, and Jain]{sleec}
Kush Bhatia, Himanshu Jain, Purushottam Kar, Manik Varma, and Prateek Jain.
\newblock Sparse local embeddings for extreme multi-label classification.
\newblock In \emph{Advances in neural information processing systems}, pages
  730--738, 2015.

\bibitem[Chen and Lin(2012)]{chen2012feature}
Yao-Nan Chen and Hsuan-Tien Lin.
\newblock Feature-aware label space dimension reduction for multi-label
  classification.
\newblock In \emph{Advances in Neural Information Processing Systems}, pages
  1529--1537, 2012.

\bibitem[Kulis and Darrell(2009)]{kulis2009learning}
Brian Kulis and Trevor Darrell.
\newblock Learning to hash with binary reconstructive embeddings.
\newblock In \emph{Advances in neural information processing systems}, pages
  1042--1050, 2009.

\bibitem[Wang et~al.(2017)Wang, Zhang, Sebe, Shen, et~al.]{wang2017survey}
Jingdong Wang, Ting Zhang, Nicu Sebe, Heng~Tao Shen, et~al.
\newblock A survey on learning to hash.
\newblock \emph{IEEE transactions on pattern analysis and machine
  intelligence}, 40\penalty0 (4):\penalty0 769--790, 2017.

\bibitem[Zamani et~al.(2018)Zamani, Dehghani, Croft, Learned-Miller, and
  Kamps]{snrm}
Hamed Zamani, Mostafa Dehghani, W~Bruce Croft, Erik Learned-Miller, and Jaap
  Kamps.
\newblock From neural re-ranking to neural ranking: Learning a sparse
  representation for inverted indexing.
\newblock In \emph{Proceedings of the 27th ACM International Conference on
  Information and Knowledge Management}, pages 497--506, 2018.

\bibitem[Guo et~al.(2019)Guo, Mousavi, Wu, Holtmann-Rice, Kale, Reddi, and
  Kumar]{glas}
Chuan Guo, Ali Mousavi, Xiang Wu, Daniel~N Holtmann-Rice, Satyen Kale, Sashank
  Reddi, and Sanjiv Kumar.
\newblock Breaking the glass ceiling for embedding-based classifiers for large
  output spaces.
\newblock In \emph{Advances in Neural Information Processing Systems}, pages
  4944--4954, 2019.

\bibitem[Gutmann and Hyv{\"a}rinen(2010)]{gutmann2010noise}
Michael Gutmann and Aapo Hyv{\"a}rinen.
\newblock Noise-contrastive estimation: A new estimation principle for
  unnormalized statistical models.
\newblock In \emph{Proceedings of the Thirteenth International Conference on
  Artificial Intelligence and Statistics}, pages 297--304, 2010.

\bibitem[Medini et~al.(2019)Medini, Huang, Wang, Mohan, and Shrivastava]{mach}
Tharun Kumar~Reddy Medini, Qixuan Huang, Yiqiu Wang, Vijai Mohan, and Anshumali
  Shrivastava.
\newblock Extreme classification in log memory using count-min sketch: A case
  study of amazon search with 50m products.
\newblock In \emph{Advances in Neural Information Processing Systems 32}, pages
  13265--13275. 2019.

\bibitem[Weinberger et~al.(2009)Weinberger, Dasgupta, Langford, Smola, and
  Attenberg]{feathash}
Kilian Weinberger, Anirban Dasgupta, John Langford, Alex Smola, and Josh
  Attenberg.
\newblock Feature hashing for large scale multitask learning.
\newblock In \emph{Proceedings of the 26th annual international conference on
  machine learning}, pages 1113--1120, 2009.

\bibitem[Zhang et~al.(2017)Zhang, Yu, Kumar, and Chang]{zhang2017learning}
Xu~Zhang, Felix~X Yu, Sanjiv Kumar, and Shih-Fu Chang.
\newblock Learning spread-out local feature descriptors.
\newblock In \emph{Proceedings of the IEEE International Conference on Computer
  Vision}, pages 4595--4603, 2017.

\bibitem[Koch(2015)]{siamese}
Gregory Koch.
\newblock Siamese neural networks for one-shot image recognition.
\newblock 2015.

\bibitem[Prabhu et~al.(2018)Prabhu, Kag, Harsola, Agrawal, and Varma]{parabel}
Yashoteja Prabhu, Anil Kag, Shrutendra Harsola, Rahul Agrawal, and Manik Varma.
\newblock Parabel: Partitioned label trees for extreme classification with
  application to dynamic search advertising.
\newblock In \emph{Proceedings of the 2018 World Wide Web Conference}, pages
  993--1002, 2018.

\bibitem[Jain et~al.(2019)Jain, Balasubramanian, Chunduri, and Varma]{slice}
Himanshu Jain, Venkatesh Balasubramanian, Bhanu Chunduri, and Manik Varma.
\newblock Slice: Scalable linear extreme classifiers trained on 100 million
  labels for related searches.
\newblock In \emph{Proceedings of the Twelfth ACM International Conference on
  Web Search and Data Mining}, 2019.

\bibitem[Varma(2014)]{xml_repo}
Manik Varma.
\newblock {Extreme Classification Repository}.
\newblock \url{http://manikvarma.org/downloads/XC/XMLRepository.html}, 2014.

\bibitem[Shrivastava and Li(2014)]{NIPS2014_5329}
Anshumali Shrivastava and Ping Li.
\newblock Asymmetric lsh (alsh) for sublinear time maximum inner product search
  (mips).
\newblock In Z.~Ghahramani, M.~Welling, C.~Cortes, N.~D. Lawrence, and K.~Q.
  Weinberger, editors, \emph{Advances in Neural Information Processing Systems
  27}, pages 2321--2329. Curran Associates, Inc., 2014.
\newblock URL
  \url{http://papers.nips.cc/paper/5329-asymmetric-lsh-alsh-for-sublinear-time-maximum-inner-product-search-mips.pdf}.

\bibitem[Charikar(2002)]{charikar2002similarity}
Moses~S Charikar.
\newblock Similarity estimation techniques from rounding algorithms.
\newblock In \emph{Proceedings of the thiry-fourth annual ACM symposium on
  Theory of computing}, pages 380--388, 2002.

\bibitem[Wang et~al.(2013)Wang, Wang, Yu, and Li]{wang2013order}
Jianfeng Wang, Jingdong Wang, Nenghai Yu, and Shipeng Li.
\newblock Order preserving hashing for approximate nearest neighbor search.
\newblock In \emph{Proceedings of the 21st ACM international conference on
  Multimedia}, pages 133--142, 2013.

\end{thebibliography}
\bibliographystyle{unsrtnat}

\appendix
\section{More for Related Work}\label{sec:lth}
\subsection{Learning to Hash}\label{subsec:lth}
Learning to Hash (LTH)~\citep{wang2017survey} has become a very integral research area for Near Neighbor Search recently. LTH learns a function $y=f(x)$ for a vector $x$ in such a way that $f$ accomplishes two goals:

1) $y$ is a compact representation of $x$

2) nearest neighbor of $x$, $x'$ has $y' = f(x')$ such that $y'$ and $y$ are as close as possible if not same.

The $2^{nd}$ condition is a restatement of the typical Locality Sensitive Hashing (LSH)~\citep{NIPS2014_5329} constraint. The major difference between LTH and LSH is the fact that $f$ is learned. 

Assume a scenario of learning $d$ dimensional vector representations for product search and using Near Neighbor Search to lookup related products. An LSH algorithm maps each vector to a compact hash code. Many similar vectors can have the same hash code (LSH property). For a lookup, we compute the hash code of a query vector and perform a true Near Neighbor Search with the candidates having the same hash code. All LSH algorithms assume the vectors to be distributed throughout the $d$-space. For example, the typical Simhash or Signed Random Projection (SRP)~\citep{charikar2002similarity} algorithm samples a few hyper-planes randomly in the $d$-space. For each label vector, it performs a dot product with all hyper-planes. For each plane, if the dot product of a label vector is $\geq 0$, it assigns a bit `1', otherwise `0'. In this way, if we sample $P$ hyper-planes, we get $P$-bit hash code for every label vector. These label vectors obey the above two properties. However, if the vectors are not distributed uniformly in the $d$ space, simhash miserably assigns a lot of vectors to the same hash code making the Near Neighbor Search almost exhaustive and slow. 

Hence, recent LTH Approaches~\citep{wang2013order} propose to optimize two objective functions: 1) pairwise similarity-preserving 2) bucket balance by maximizing entropy of learned representations. However, no objective function has successfully solved the problem of imbalanced Hash functions.

SOLAR tackles this problem in a unique way by assigning load uniformly to each bucket and then repeating this process multiple times to ensure the distinguishability of dissimilar items. So our hash tables are predetermined and fixed. We learn to map the inputs to respective buckets which might return relevant candidates along with several irrelevant ones too. However, the final step of scoring and sorting eliminates the irrelevant ones and accomplishes the overall functionality of a hash table.

\subsection{SNRM}\label{subsec:snrm}
While there have been a plethora of dense embedding models, there is only one prior work called SNRM (Standalone Neural Ranking Model)~\citep{snrm} that trains sparse embeddings for the task of suggesting documents relevant to an input query (classic web search problem). In SNRM, the authors propose to learn a high dimensional output layer and sparsify it using a typical L1 or L2 regularizer. Further, they opt for a weak-labelling strategy wherein each training sample has a query $q$, documents $d1$ and $d2$ and a binary label $l$ suggesting whether $d1$ is more relevant to $q$ than $d2$ or vice-versa. The query and documents are passed through a Neural Network (RNN for a text input) to obtain their high-dimensional sparse embeddings. These embeddings are concatenated and a hinge-loss is trained to learn the label $l$. After training, a standard inverted-index is constructed where all the products are partitioned based on the non-zero indexes in their sparse embeddings. During inference, a query is mapped to its sparse vector using the neural network and the non-zero indexes are obtained. All documents that have atleast one matching non-zero index with the query are shortlisted and scored to get the best predictions.

While SNRM looks good to be an efficient sparse alternative to dense embeddings, imposing sparsity through regularization has multiple issues - 1) The training and inference becomes too sensitive to the regularization weight. A larger regularization weight causes the embeddings to be too sparse and in all likelihood, we retrieve zero labels for many inputs. On the other hand, if the regularization weight is very small, we end up retrieving too many candidates defeating the purpose of sparse embeddings. 2) The inverted-index generally has a lopsided label distribution causing imbalanced loads and high inference times. As we see in our experiments later, these issues lead to the poor performance of SNRM on our 1.67M product recommendation dataset.

\section{Positive-only Association}\label{sec:pos_only}
The random choice of buckets in constructing SOLAR embedding might pool totally unrelated labels into a bucket in each component, For example, two books titled `Velveteen Rabbit' and `Gravitational Waves' might be assigned the same bucket in a particular component of the embedding. Even though this pooling appears unconventional, it is necessary to maintain the load balance of each bucket. However, the convergence of cross-entropy loss might come under the scanner in such a scenario.

Going by our training design, we choose to train that specific bucket with `1' if the input query is either related to `Velveteen Rabbit' or `Gravitational Waves'. By choosing an `OR' operation over the true labels, we enforce the model to reasonably fire up the bucket for all queries containing terms `waves', `gravity', `rabbit' etc. This is called `positive-only association'. Although a single bucket might correspond to unrelated labels, the labels that have high scores in all of the $K$ components would be truly related to the input.

\section{Law of Deferred Orthogonality of Learned Embedding}\label{sec:def_orthogonality}
\begin{lemma}\label{lemma:1}
For any two orthonormal basis matrices, $\bf A = [\overline{\bf a_1}\ \overline{\bf a_2}\ \overline{\bf a_3}\ ...\ \overline{\bf a_n}]$ and $\bf B = [\overline{\bf b_1}\ \overline{\bf b_2}\ \overline{\bf b_3}\ ...\ \overline{\bf b_n}]$ in the same vector space, there exists an orthogonal matrix $\bf P$ such that $\bf A = \bf P\bf B$.
\end{lemma}
Note that, since we construct fixed norm label vectors, orthogonal vectors form orthonormal basis vectors (up to a constant norm factor multiplication).

\begin{proof} Proving the existence is simple. We can perform an orthogonal projection of $a_i$ on to $B$ as:
$$a_i = (a_i\cdot b_1)b_1 +...+(a_i\cdot b_n)b_n\  = \sum_j b_j(a_i\cdot b_j)$$
Hence, a matrix $P$ can be designed as $P = (P_i^j) = (a_i\cdot b_j)$ to obey $A = PB$. For the orthogonality, consider $$(PP^T)_i^j = \sum_k (b_j\cdot a_k)(a_k\cdot b_i) = \sum_k (b_j^T a_K)(a_k^T b_i) = b_j^T \left(\sum_k(a_K a_k^T)\right)b_i$$
Since $A$ and $B$ have orthonormal columns, we have $\sum_k(a_K a_k^T) = I$ and $(PP^T)_i^j = b_j^Tb_i = \delta_{ji}$.
This leads to the orthogonality of $P$.
$$PP^T = I\implies P^T = P^{-1}$$ 
\end{proof}

\begin{theorem} {\bf Law of Deferred Orthogonality of Learned Embedding}
Given any positive semi-definite kernel $S(.,.)$, and functions $f_I$ and $f_L$, where $f_L$ is orthogonal. For any orthogonal function $R$ with the same input and range space as $f_L$, there always exist a function $f$, such that
$S(f_I(x),f_L(y)) = S(f(x),R(y)).$
\label{thm:orthogonal}
\end{theorem}

\begin{proof}
Lemma \ref{lemma:1} states that we can transform one orthonormal basis to another using an orthogonal matrix. Let the SOLAR's label vectors be denoted by ${\bf A}= [ \overline{ {\bf R}(y_1)},\ \overline{{\bf R}(y_2)},\ ...,\  \overline{{\bf R}(y_N)} ]$. Here, $\bf R$ refers to the random initialization matrix that maps label ids to sparse vectors. Similarly, the two-sided learnt label vectors are denoted by ${\bf B} = [ \overline{f_L(y_1)},\ \overline{f_L(y_2)},\ ...,\ \overline{f_L(y_N)} ]$. Let $\bf P$ be the transformation matrix from $\bf B$ to $\bf A$, i.e., $A_i = R(y_i) = {\bf P} B_i = {\bf P} f_L(y_i)$. By virtue of orthogonality, $\bf P$ preserves the inner product objective function as follows:
$$\langle f_I(x),f_L(y) \rangle \ =\ \langle {\bf P}f_I(x),{\bf P}f_L(y)\rangle\ =\ \langle {\bf P}f_I(x), {\bf R}(y)\rangle$$

Hence learning one function $f_{SOLAR}\ =\ {\bf P}f_I$ is equivalent to learning two function $f_I$ and $f_L$ provided the columns of ${\bf B}$ are orthonormal.

\end{proof}

\begin{table*}[t]
\hspace{-2.7cm}
\begin{tabular}{ | p{6cm} | p{6cm} | p{6cm} |}
\hline
{\bf Query 1:} remington the science and practice of pharmacy 2 volumes & {\bf Query 2:} hello in there a big sisters book of waiting  & {\bf Query 3:} beginners guide to american mah jongg how to play the game amp win \\
\hline
\hline
{\bf SOLAR Top Preds (All Correct)} & {\bf SOLAR Top Preds (All Correct)} & {\bf SOLAR Top Preds (only 3 correct)}\\
\hline
\hline
pharmacotherapy a patho-physiologic approach 8th edition & im a big sister & winning american mah jongg strategies a guide for the novice player\\
\hline
the sanford guide to antimicrobial therapy sanford guides & you and me new baby & fudge dice black 4 dice in plastic tube\\
\hline
goodman and gilmans the pharmacological basis of therapeutics twelfth edition & look at me new baby & the great mahjong book history lore and play\\
\hline
basic and clinical pharmacology 12e lange basic science & the berenstain bears new baby & beginners guide to american mah jongg how to play the game amp win\\
\hline
\hline
{\bf GLaS Top Preds (All Correct)} & {\bf GLaS Top Preds (All Wrong)} & {\bf GLaS Top Preds (All Correct)}\\
\hline
\hline
goodman and gilmans the pharmacological basis of therapeutics twelfth edition & just me and my little brother little critter picturebackr & winning american mah jongg strategies a guide for the novice player\\
\hline
pharmaceutical calculations 13th edition & the day the crayons quit & the great mahjong book history lore and play\\
\hline
basic and clinical pharmacology 12e lange basic science & the invisible boy & beginners guide to american mah jongg how to play the game amp win\\
\hline
the sanford guide to antimicrobial therapy 2013 & the name jar & national mah jongg league scorecard large 2014\\
\hline
\end{tabular}

\caption{Sample queries and the respective top predictions from SOLAR and DSSM+GLaS. The first query is a relatively frequent one. The second and third are relatively infrequent. We can see that on infrequent queries, SOLAR is more robust than dense embedding models.}

\label{tab:samples}
\end{table*}

\section{Additional Information for Experiments}\label{sec:addn_exp}

{\bf Baselines and our settings:} The most natural comparison arises with recent industry standard embedding models, namely Amazon's Deep Semantic Search Model (DSSM)~\citep{dssm} and Facebook's Deep Learned Recommendation Model (DLRM)~\citep{dlrm}. DLRM takes mixed inputs (dense and sparse) and learns token embeddings through a binary classifier for ad prediction. On the other hand, DSSM trains embeddings for semantic matching where we need to suggest relevant products for a user typed query. Because DSSM's goal aligns with this dataset, we compare SOLAR against it. For DSSM, we train a $763265 \times 1600$ embedding matrix where each word in the vocabulary has a 1600 dimensional dense vector (to be consistent with the query sparsity of SOLAR). This matrix is shared across input and output titles (Siamese Network). We tokenize a title into words and lookup their embeddings and mean-pool them. This is followed by a batch normalization and $tanh$ activation. The resultant embeddings of both input and label titles are optimized to have high cosine similarity. For every input title, we also train to minimize cosine similarity with one irrelevant label title (Negative Sampling).

We additionally incorporate orthogonality on DSSM label embeddings using the recent GLaS~\citep{glas} regularizer. We shuffle the entire training data, pick one label from each row and obtain the label vector matrix ${\bf V}$. Then, we add a $2^{nd}$ optimization function (in addition to the cosine similarity) of the form:
\begin{equation}
l_{GLaS} = \lambda \Vert {\bf V}^T{\bf V} - \frac{1}{2}({\bf CZ}^{-1}\ +\ {\bf Z}^{-1}{\bf C})\Vert_2^2
\end{equation} 
where, ${\bf C}$ is the label co-occurrence matrix and $Z$ is the diagonal component of ${\bf C}$. Hence, ${\bf CZ}^{-1}[i,j] = p(i|j)$ and ${\bf Z}^{-1}{\bf A}[i,j] = p(j|i)$.

Another natural comparison arises with the lone sparse embedding model SNRM~\citep{snrm}. We did try to use the available code for SNRM with our data generator. However, for reasons explained in section \ref{subsec:snrm}, we could not get any reasonable metrics even with very low regularization weight. The training would eventually culminate in the learned vectors being absolutely zero and thereby retrieving nothing from the inverted indexes. Even the label vectors end up being empty most of the time. We included SNRM results in the main paper.

{\bf Sample queries and predictions for SOLAR vs DSSM:} We perform a qualitative assessment of SOLAR and DSSM+GLaS to assess the robustness to infrequent and spurious queries. For each test query, we sum up the frequency of each term in the corpus and sort the queries. We pick a random query from the top half and classify it as a `frequent query'. Similarly, we pick a random query from the bottom half and classify it as an `infrequent query'. We manually examined 10 such queries and listed three of them in table \ref{tab:samples}. The first query was classified as frequent and the rest two as infrequent. We show the top 4 predictions from either algorithm and the number of ground truth labels among them.

\end{document}